\documentclass[10pt,twocolumn,letterpaper]{article}

\usepackage{iccv}
\usepackage{times}
\usepackage{epsfig}
\usepackage{graphicx}
\usepackage{amsmath}
\usepackage{amssymb}
\usepackage[font={small}]{caption}
\usepackage{multirow}
\usepackage{array}
\usepackage{setspace}

\usepackage[pagebackref=true,breaklinks=true,colorlinks,letterpaper=true,bookmarks=false]{hyperref}

\newenvironment{packed_itemize}{
	\begin{itemize}
		\setlength{\itemsep}{-0.5pt}
		\setlength{\parskip}{-0.5pt}
		\setlength{\parsep}{-5pt}
	}{\end{itemize}}

\usepackage{lipsum}

\iccvfinalcopy 

\def\iccvPaperID{2035} 

\ificcvfinal\fi

\begin{document}

\title{Unsupervised Domain Adaptive 3D Detection with Multi-Level Consistency}


\author{Zhipeng Luo$^{1,2*}$ \;
Zhongang Cai$^{1,2,3}$\thanks{Equal contribution} \;
Changqing Zhou$^{2,4*}$ \;
Gongjie Zhang$^4$ \;
Haiyu Zhao$^{2,3}$ \; \\
Shuai Yi$^{2,3}$ \;
Shijian Lu$^{4}$\thanks{Corresponding author} \;
Hongsheng Li$^5$ \;
Shanghang Zhang$^6$ \;
Ziwei Liu$^1$  \; \\
$^1$ S-Lab, Nanyang Technological University \quad
$^2$ Sensetime Research \quad
$^3$ Shanghai AI Laboratory \\ 
$^4$ Nanyang Technological University \quad
$^5$ Chinese University of Hong Kong \quad
$^6$ UC Berkeley \quad \\
{\tt\small \{zhipeng001, zhou0365\}@e.ntu.edu.sg, \{caizhongang, zhaohaiyu, yishuai\}@sensetime.com} \\
{\tt\small \{shijian.lu\}@ntu.edu.sg, hsli@ee.cuhk.edu.hk, shz@eecs.berkeley.edu, zwliu.hust@gmail.com} 
}


\makeatletter
\newcommand{\thickhline}{%
    \noalign {\ifnum 0=`}\fi \hrule height 1pt
    \futurelet \reserved@a \@xhline
}
\newcolumntype{"}{@{\hskip\tabcolsep\vrule width 1pt\hskip\tabcolsep}}
\makeatother

\newcommand{\name}{MLC-Net}

\maketitle

\ificcvfinal\thispagestyle{empty}\fi


\begin{abstract}

Deep learning-based 3D object detection has achieved unprecedented success with the advent of large-scale autonomous driving datasets. 
However, drastic performance degradation remains a critical challenge for cross-domain deployment. 
In addition, existing 3D domain adaptive detection methods often assume prior access to the target domain annotations, which is rarely feasible in the real world. To address this challenge, we study a more realistic setting, unsupervised 3D domain adaptive detection, which only utilizes source domain annotations.
\textbf{1)} We first comprehensively investigate the major underlying factors of the domain gap in 3D detection. Our key insight is that geometric mismatch is the key factor of domain shift. 
\textbf{2)} Then, we propose a novel and unified framework, \textbf{Multi-Level Consistency Network (\name{})}, which employs a teacher-student paradigm to generate adaptive and reliable pseudo-targets.
\name{} exploits point-, instance- and neural statistics-level consistency to facilitate cross-domain transfer. 
Extensive experiments demonstrate that \name{} outperforms existing state-of-the-art methods (including those using additional target domain information) on standard benchmarks. Notably, our approach is detector-agnostic, which achieves consistent gains on both single- and two-stage 3D detectors. Code will be released.

\end{abstract}

\begin{figure}[t]
    \centering
    \includegraphics[width=1.0\linewidth]{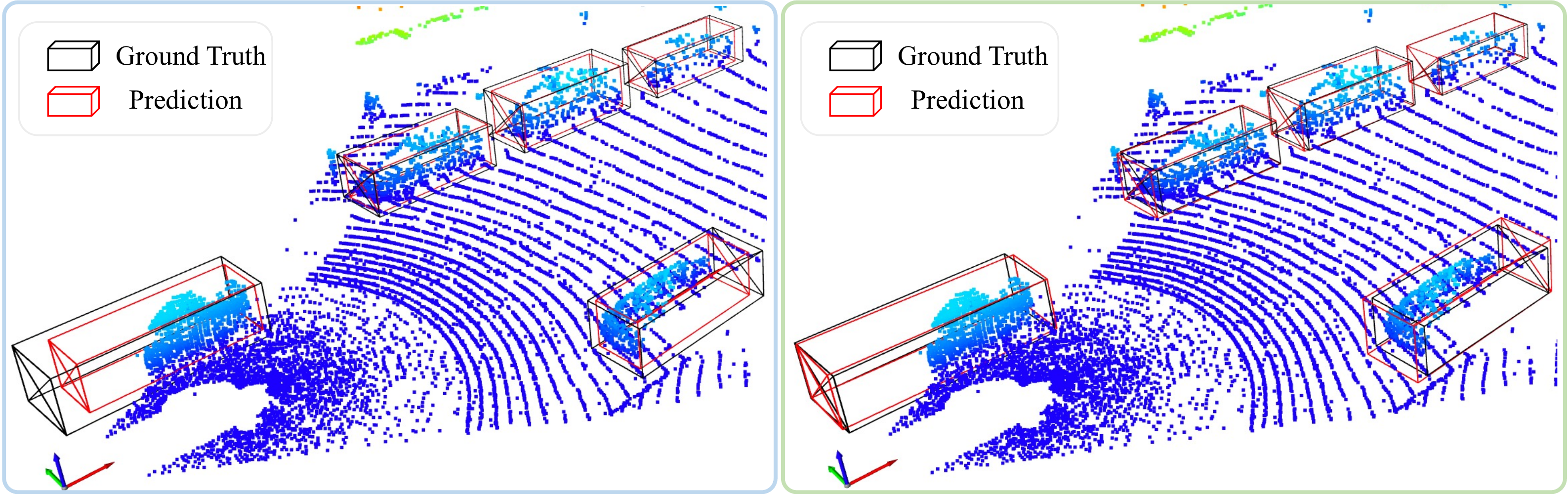}
    \caption{Visualization of detection results for domain adaptation from KITTI to Waymo dataset. \textbf{Left}: Predictions of baseline model trained on KITTI dataset and directly tested on Waymo dataset. The model can classify and localize the objects, but produces inaccurate box scale due to geometric mismatch. The predicted boxes are therefore noticeably smaller than the ground truth. \textbf{Right}: Predictions of our domain-adaptive \name{}, which demonstrates accurate bounding box scale even though \name{} is trained without access to any target domain annotations. Best viewed in color.}
    \label{fig:teaser}
    \vspace{-0.5em}
\end{figure}

\section{Introduction}
\label{sec:intro}

\begin{figure*}[t]
    \centering
    \includegraphics[width=1.0\linewidth]{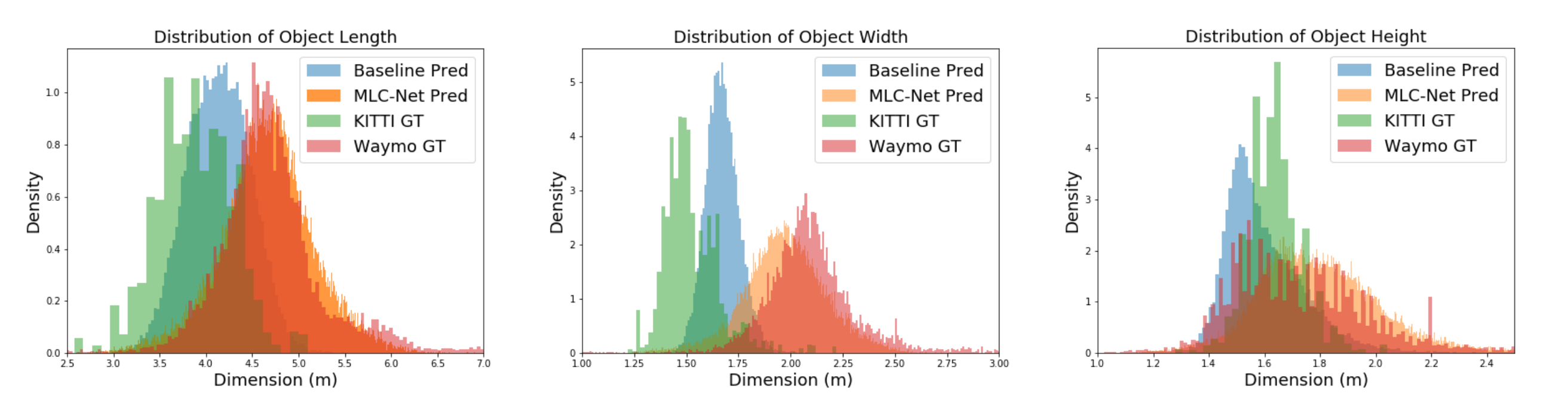}
    \caption{A study on the domain shift for 3D detection. Here we take KITTI as the source dataset and Waymo as the target dataset. Our key insights include: 1) distribution of object dimensions varies drastically across datasets, indicating geometric mismatch can be a key factor for the domain gap; 2) directly applying a model trained on KITTI to Waymo (referred to as the baseline in the figure) is ineffective: the model continues to predict box dimensions close to the source domain; 3) our \name{} is effective in addressing the geometric mismatch, and the distributions of its predictions on the target domain accurately align with the ground truth. Best viewed in color.}
    \label{fig:distribution}
    \vspace{-0.5em}
\end{figure*}

With the prevalent use of LiDARs for autonomous vehicles and mobile robots, 3D object detection on point clouds has drawn increasing research attention. Large-scale 3D object detection datasets \cite{geiger2013kittidataset, sun2020waymo, caesar2020nuscenes} in recent years has empowered deep learning-based models \cite{shi2019pointrcnn, yang20203dssd, yan2018second, lang2019pointpillars, shi2020pvrcnn, yang2019std, qi2019votenet, sindagi2019mvx, shi2020parta2, zhu2020ssn, yin2020centerpoint} to achieve remarkable success. However, deep learning models trained on one dataset (source domain) often suffer tremendous performance degradation when evaluated on another dataset (target domain). We investigate the bounding box scale mismatch problem (e.g., vehicle size in the U.S. is noticeably larger than that in Germany), which is found to be a major contributor to the domain gap, in alignment with previous work~\cite{wang2020traininger}. This is unique to 3D detection: compared to 2D bounding boxes that can have a large variety of size, depending on the distance of the object from the camera, 3D bounding boxes have a more consistent size in the same dataset, regardless of the objects' location relative to the LiDAR sensor. Hence, the detector tends to memorize a narrow and dataset-specific distribution of bounding box size from the source domain (Figure \ref{fig:distribution}).

Unfortunately, existing works are inadequate to address the domain gap with a realistic setup. Recent methods on domain adaptive 3D detection either require some labeled data from the target domain for finetuning or utilize some additional statistics (such as the mean size) of the target domain \cite{wang2020traininger}. However, such knowledge of the target domain is not always available. In addition, popular 2D unsupervised domain adaptation methods that leverage feature alignment techniques \cite{chen2018dafastercnn, saito2019strongweak, zheng2020crossdomain, guan2021uncertainty, chen2020harmonizing, guan2021scale,xu2020exploring, huang2020contextual, li2020spatialattention, huang2021fsdr, zhan2019ga, huang2021rda, zhao2021domain, tseng2020cross} to mitigate domain shift are not readily transferable to 3D detection. While these methods are effective in handling domain gaps due to lighting, color, and texture variations, such information is unavailable in point clouds. Instead, point clouds pose unique challenges such as the geometric mismatch discussed above.

Therefore, we propose \name{} for unsupervised domain adaptive 3D detection. \name{} is designed to tackle two major challenges. First, to create meaningful scale-adaptive targets to facilitate the learning, \name{} employs the mean teacher \cite{tarvainen2017meanteacher} learning paradigm. The teacher model is essentially a temporal ensemble of student models: the parameters of the teacher model are updated by an exponential moving average window on student models of preceding iterations. Our analyses show that the mean teacher produces accurate and stable supervision for the student model without any prior knowledge of the target domain. To the best of our knowledge, we are the first to introduce the mean teacher paradigm in unsupervised domain adaptive 3D detection. Second, to design scale-related consistency losses and construct useful correspondences of teacher-student predictions to initiate gradient flow, we design \name{} to enforce consistency at three levels. \textbf{1)} Point-level. As point clouds are unstructured, point-based region proposals or equivalents \cite{shi2019pointrcnn, yang20203dssd} are common. Hence, we sample the same subset of points and share them between the teacher and student. We retain the indices of the points that allow 3D augmentation methods to be applied without losing the correspondences. \textbf{2)} Instance-level. Matching region proposals can be erroneous, especially at the initial stage when the quality of region proposals is substandard. Hence, we resort to transferring teacher region proposals to students to circumvent the matching process. \textbf{3)} Neural statistics-level. As the teacher model only accesses the target domain input, the mismatch between the batch statistics hinders effective learning. We thus transfer the student's statistics, which is gathered from both the source and the target domain, to the teacher to achieve a more stable training behavior.

\name{} shows remarkable compatibility with popular mainstream 3D detectors, allowing us to implement it on both two-stage~\cite{shi2019pointrcnn} and single-stage~\cite{yang20203dssd} detectors. Moreover, we verify our design through rigorous experiments across multiple widely used 3D object detection datasets \cite{geiger2013kittidataset, sun2020waymo, caesar2020nuscenes}. Our method outperforms baselines by convincing margins, even surprisingly surpassing existing methods that utilize additional information. In summary, our main \textbf{contributions} are:
\begin{packed_itemize}

    \item We formulate and study unsupervised domain adaptive 3D detection, a pragmatic, yet underexplored task that requires no annotations of the target domain. We comprehensively investigate the major underlying factors of the domain gap in 3D detection and find geometric mismatch is the key factor. 
    \item We propose a concise yet effective mean-teacher paradigm that leverages three levels of consistency to facilitate cross-domain transfer, achieving a significant performance boost that is consistent across multiple popular public datasets.
    \item We validate our hypothesis on the unique challenges associated with point clouds and verify our proposed approach with comprehensive evaluations, which we hope would lay a strong foundation for future research.

\end{packed_itemize}

\section{Related Works}
\label{sec:related_works}
\noindent\textbf{LiDAR-based 3D Detection.}
LiDAR-based 3D detection methods mainly come from two categories, namely grid-based methods and point-based methods. Grid-based approaches convert the whole point cloud scene to voxels of fixed size and process the input with 2D or 3D CNN. MV3D \cite{mv3d} first projects point clouds to bird-eye view images to generate proposals. PointPilar \cite{lang2019pointpillars} performs voxelization on point clouds and converts the representation to 2D. VoxelNet \cite{zhou2018voxelnet} obtains voxel representations by applying PointNet \cite{qi2017pointnet} to points and processes the features with 3D convolution. SECOND \cite{yan2018second} applies 3D sparse convolution \cite{graham20183dsparseconv} to improve the efficiency. PV-RCNN \cite{shi2020pvrcnn} proposes to combine voxelization and point-based set abstraction to obtain more discriminative features. On the other hand, point-based methods directly extract features from raw point cloud input. F-PointNet \cite{qi2018frustum} applies PointNet \cite{qi2017pointnet} to perform 3D detection based on 2D bounding boxes. PointRCNN \cite{shi2019pointrcnn} proposes a two-stage framework to generate box bounding proposals from the whole point clouds and refine them with feature pooling. 3DSSD \cite{yang20203dssd} proposes to use F-FPS for better point sampling to achieve single-stage detection. In this work, we conduct focused discussion with PointRCNN as the 
base model but we show our method is also compatible to single-stage detector (3DSSD) in Supplementary Material.

\smallskip
\noindent\textbf{Point Cloud Domain Adaptation.}
While extensive researches have been conducted on domain adaptation tasks with 2D image data, the 3D point cloud domain adaptation field has relatively small literature. PointDAN \cite{qin2019pointdan} proposes to jointly align local and global features using discrepancy loss and adversarial training for point cloud classification. Achituve et. al. \cite{achituve2021selfsuupervised} introduces an additional self-supervised reconstruction task to improve the classification performance on the target domain. Yi et. al. \cite{yi2020completeandlabel} designs a sparse voxel completion network to perform point cloud completion for domain adaptive semantic segmentation. Jaritz et. al. \cite{jaritz2020xmuda} leverages multi-modal information by projecting point cloud to 2D images and train models jointly. For object detection, \cite{wang2020traininger} identifies the major domain gap of object size mismatch among autonomous driving datasets and proposes to mitigate the gap by leveraging target domain object scale statistics. SF-UDA \cite{saltori2020sourcefree} computes motion coherence over consecutive frames to select the best scale for the target domain. Our proposed method works under a similar setup to \cite{wang2020traininger} but does not require target domain geometric statistics.

\smallskip
\noindent\textbf{Mean Teacher.} 
The mean teacher framework \cite{tarvainen2017meanteacher} is first proposed for semi-supervised learning. Many variants \cite{cubuk2018autoaugment, berthelot2019mixmatch, xie2019unsupervised} have been proposed to further improve its performance. Furthermore, the framework has also been applied to other fields such as domain adaptation \cite{french2017selfensembling, cai2019exploringobjectrelation} and self-supervised learning \cite{he2020moco, grill2020byol, liu2020selfemd} where labeled data is scarce or unavailable. Specifically, the mean teacher framework incorporates one trainable student model and a non-trainable teacher model whose weights are obtained from the exponential moving average of the student model's weights. The student model is optimized based on the consistency loss between the student and teacher network predictions. In particular, although \cite{cai2019exploringobjectrelation} also employs the mean teacher paradigm for the detection task by aligning region-level features, point cloud detection models are substantially different from 2D detectors and our proposed method differs by incorporating multi-level consistency.

\begin{figure*}[t] 
    \centering
    \includegraphics[width=1.0\linewidth]{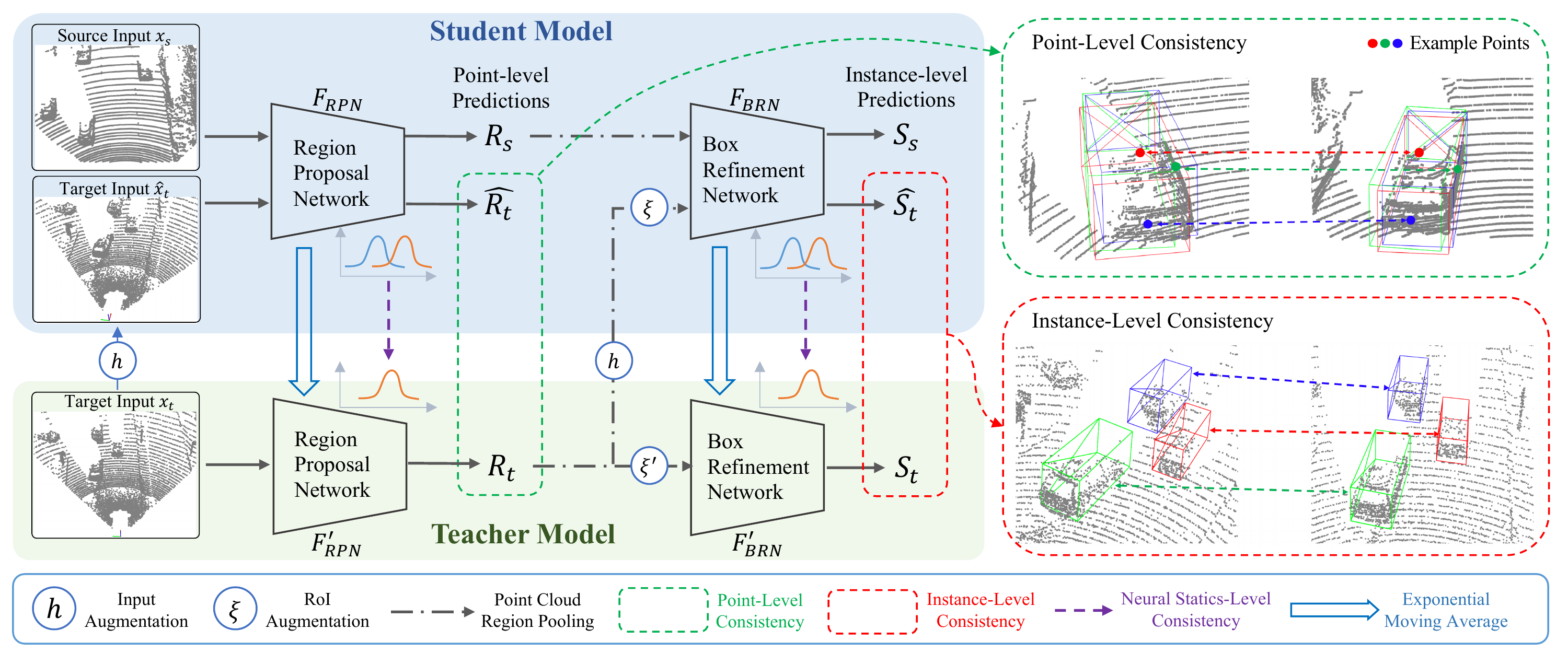}
    \caption{The network architecture of our proposed \name{}. \name{} leverages the mean-teacher \cite{tarvainen2017meanteacher} paradigm where the teacher is the exponential moving average (hence the name mean-teacher) of the student model and is updated at every iteration. This mean-teacher design provides high-quality pseudo labels to facilitate smooth learning of the student model. Towards the goal, we design consistency enforced at three levels. First, at point-level, 3D proposals are associated based on point correspondences, which are established by sampling the same set of points from the target domain for both the student and teacher models; second, at  instance-level, the teacher 3D proposals are passed to the student Box Refinement Network, and the correspondences between 3D box predictions from two models are naturally maintained; third, at neural statistics-level, we discover non-learnable parameters in batch normalization layers demonstrate significant domain shift, and thus align the teacher's parameters with the student's. We highlight the efficacy of \name{} and further discuss our design motivations in Section \ref{sec:method}. Best viewed in color.}    
    \label{fig:model}
\end{figure*}

\section{Our Approach}
\label{sec:method}
In this section, we formulate the 3D point cloud domain adaptive detection problem (Section \ref{subsec:problem_definition}), and provide an overview of our \name{} (Section \ref{subsec:overview}), followed by the details of our mean-teacher paradigm (Section \ref{subsec:mean_teacher}). Finally, we explain the details of the point-level (Section \ref{subsec:point-level_consistency}), instance-level (Section \ref{subsec:instance-level_consistency}), and statistics-level (Section \ref{subsec:statistics-level_consistency}) consistency of our \name{}.

\subsection{Problem Definition}
\label{subsec:problem_definition}
Under the unsupervised domain adaptation setting, we have access to point cloud data from one labeled source domain $\mathbb{D}_s = {\{x_s^i, y_s^i\}}_{i=1}^{N_s}$ and one unlabeled target domain $\mathbb{D}_t = {\{x_t^i\}}_{i=1}^{N_t}$, where $N_s$ and $N_t$ are the number of samples from the source and target domains, respectively. Each point cloud scene $x^i\in\mathbb{R}^{n\times3}$ consists of $n$ points with their 3D coordinates while $y^i$ denotes the label of the corresponding training sample from the source domain. $y$ is in the form of object class $k$ and 3D bounding box parameterized by the center location of the bounding box $(c_x, c_y, c_z)$, the size in each dimension $(d_x, d_y, d_z)$, and the orientation $\eta$. The goal of the domain adaptive detection task is to train a model $F$ based on $\mathbb{D}_s$ and $\mathbb{D}_t$ and maximize the performance on $\mathbb{D}_t$.

\subsection{Framework Overview}
\label{subsec:overview}
We illustrate \name{} in Figure \ref{fig:model}. The labeled source input $x_{s}$ is used for standard supervised training of the student model $F$ with loss $L_{source}$. For each unlabeled target domain example $x_t$, we perturb it by applying random augmentation $h$ to obtain $\hat{x_t}$. The perturbed and original point cloud inputs are passed to the student model and teacher model respectively to get their point-level box proposals $\hat{R_t}$ and $R_t$ where point-level consistency is applied. Subsequently, teacher proposals are augmented with $h$ and passed to the student model for box refinement, to obtain $\hat{S_t}$. Together with teacher's instance-level predictions $S_t$, the instance-level consistency is applied. The overall consistency loss $L_{consist}$ is computed as:
\begin{equation}
    L_{consist} = L_{pt, cls} + L_{pt, box} + L_{ins, cls} + L_{ins, box}
\end{equation}
where pt, ins, cls and box stand for point-level, instance-level, classification and box regression respectively. These loss components are elaborated in Section~\ref{subsec:point-level_consistency} and ~\ref{subsec:instance-level_consistency}. In each iteration, the student model is updated through gradient descent with the total loss ${L}$, which is a weighted sum of $L_{source}$ and $L_{consist}$:
\begin{equation}
    L = \lambda L_{source} + L_{consist}
\end{equation}
where $\lambda$ is the weight coefficient. The learnable parameters of the student model are then used to update the corresponding teacher model parameters, where the details can be found in Section \ref{subsec:mean_teacher}. In addition, we enforce non-learnable parameters to be aligned between the teacher and the student via neural statistics-consistency (Section~\ref{subsec:statistics-level_consistency}).

\name{} achieves two major design goals towards effective unsupervised 3D domain adaptive detection. \textbf{First}, to generate accurate and robust pseudo targets without any access to the target domain annotation or statistical information. \name{} leverages a mean teacher paradigm where the teacher model can be regarded as a temporal ensemble of student models, allowing it to produce high-quality predictions and guide the learning of the student. \textbf{Second}, to design effective consistency losses at point-, instance- and neural statistics-level that enhance adaptability to scale variation, and construct the teacher-student correspondences that allow the back-propagated gradient to flow through the correct routes. Although we conduct most analysis on PointRCNN as the representative of two-stage 3D detectors, we highlight that our method is generic and can be easily extended to single-stage detection models such as 3DSSD with modest modifications (see Supplementary Material). 

\subsection{Mean Teacher}
\label{subsec:mean_teacher}
Motivated by the success of the mean teacher paradigm \cite{tarvainen2017meanteacher} in semi-supervised learning and self-supervised learning, we apply it to our point cloud domain adaptive detection task as illustrated in Figure \ref{fig:model}. The framework consists of a student model $F$ and a teacher model $F'$ with the same network architecture but different weights $\theta$ and $\theta'$ , respectively. The weights of the teacher model are updated by taking the exponential moving average of the student model weights:
\begin{equation}
    \theta' = m \theta' + (1-m) \theta
\end{equation}
where $m$ is known as the momentum which is usually a number close to 1, e.g. 0.99. 
Figure~\ref{fig:student_vs_teacher} shows that the teacher model constantly provides effective supervision to the student model via high-quality pseudo targets. Hence, by enforcing the consistency between the student and the teacher, the student learns domain-invariant representations to adapt to the unlabeled target domain, guided by the pseudo labels. We show in Table~\ref{tab:meanteacher} that the mean teacher significantly improves model performance compared to baseline.

\subsection{Point-Level Consistency}
\label{subsec:point-level_consistency}
The point-level consistency loss is calculated between the first-stage box proposals of the student and teacher models. One of the key challenges for formulating consistency is to find the correspondence between the student and the teacher. Unlike image pixels that are arranged in regular lattices, points reside in continuous 3D space which lacks structure \cite{qi2017pointnet}. Hence, constructing point correspondences can be problematic (Table \ref{tab:matching_methods}). Instead, we circumvent the difficulty by feeding the teacher and the student two identical sets of points at the very beginning and trace the point indices to maintain correspondences. 

Specifically, for each target domain example, we sample $M$ points from the point cloud scene to obtain the teacher input $x_t$ and apply random augmentation $h$ on a replicated set to obtain $\hat{x_t}$ with $\hat{x_t}=h(x_t)$. $h$ consists of random global scaling of the point cloud scenes and can be regarded as applying displacements on individual points, without disrupting the point correspondences. As a result, each point $p\in x_t$ corresponds to a point $\hat{p}\in\hat{x_t}$, and this relationship holds for the point-level predictions of the region proposal network $F_{RPN}$. We denote the first stage prediction as $R=F_{RPN}(x)$. Note that the point correspondences are transferred to box proposals as each point generates one box proposal. $R$ consists of class prediction $R^c$ and box regression $R^b$. For the class predictions, we define the consistency loss as the Kullback-Leibler (KL) divergence between each point pair from $x_t$ and $\hat{x_t}$:
\begin{equation}
    L_{pt,cls} = \frac{1}{|x_t|} \sum{D_{KL}(\hat{R}_t^c||R_t^c)}
\end{equation}
where $|x_t|$ stands for the number of points in $x_t$. 

More importantly, we enforce consistency between bounding box regression predictions to address geometric mismatch. For the bounding box predictions, we only compute the consistency over points belonging to the objects because the background points do not generate meaningful bounding boxes. We obtain a set of points $\mathbb{P}_{pos}$ which fall inside the bounding boxes of the final predictions of both the student and teacher networks with $\mathbb{P}_{pos}=\{(p\in NMS(\hat{S_t}))\cap (p\in NMS(S_t))\}$, where $\hat{S_t}$ and $S_t$ are the refined bounding box predictions after second stage (see Section~\ref{subsec:instance-level_consistency}). We then compute the point-level box consistency loss as:
\begin{equation}
    L_{pt,box} = \frac{1}{|\mathbb{P}_{pos}|} \sum_{p^i\in \mathbb{P}_{pos}}{d(\hat{R}_t^{c(i)}, h(R_t^{c(i)}))}
\end{equation}
where $d$ is the smooth $L1$ loss and $h$ is the random augmentation applied to the input $x_t$. We apply the same augmentation to the teacher bounding box predictions to align with the scale of the student point cloud scene before computing the consistency.

\subsection{Instance-Level Consistency}
\label{subsec:instance-level_consistency}
In the second stage, NMS is performed on $R$ to obtain $N$ high-confidence region proposals denoted as $G$ for each point cloud scene. We highlight that the association between region proposals from the student and teacher models are lost in the NMS process due to the differences between $\hat{R_t}$ and $R_t$. To match the instance-level predictions for consistency computation, a common method is to perform greedy matching based on IoU between teacher and student region proposals. However, such matching is not robust due to the large number of noisy predictions, which leads to ineffective learning as shown experimentally in Table \ref{tab:matching_methods}. Hence, we adopt a simple approach by replicating the teacher region proposals to the student model and applying the input augmentation $h$ to match the scale of the student model. Subsequently, we disturb the region proposals by applying random RoI augmentation $\xi$ for the sets of region proposals before they are used for feature pooling. The motivation of this operation is to force the models to output consistent predictions given non-identical region proposals and prevent convergence to trivial solutions. Formally, the above process can be described as $\boldsymbol{\hat{f_t}}=pool(\xi(h(G_t)))$ and $\boldsymbol{f_t}=pool(\xi'(G_t))$ for the student and teacher models, respectively, where $\boldsymbol{f}$ denotes the instance-level features obtained from feature pooling as described in \cite{shi2019pointrcnn}. The pooled features are then passed to the box refinement network $F_{BRN}$ for box refinement to obtain the second stage predictions $S=F_{BRN}(\boldsymbol{f})$. 
Similar to the first stage prediction $R$, $S$ consists of the class prediction $S^c$ as well as the bounding box prediction $S^b$. We define the instance-level class consistency as the difference between $\hat{S}_t^c$ and $S_t^c$:
\begin{equation}
    L_{ins,cls} = \frac{1}{|G_t|} \sum{D_{KL}(\hat{S}_t^c||S_t^c)}
\end{equation}
where $|G_t|$ denotes the number of region proposals. On the other hand, to compute the instance-level box consistency loss, we first obtain a set of positive predictions $\mathbb{S}_{pos} = \{({\hat{S}_t^c>\varepsilon) \cap (S_t^c>\varepsilon})\}$ by selecting bounding boxes with classification predictions larger than a probability threshold $\varepsilon$. We then apply $h$ to $S_t^b$ to match the scale and compute the instance-level box consistency loss based on the discrepancy between $\hat{S}_t^b$ and $S_t^b$ for the selected predictions:
\begin{equation}
    L_{ins,box} = \frac{1}{|\mathbb{S}_{pos}|} \sum_{S_t^{i} \in \mathbb{S}_{pos}}{d(\hat{S}_t^{b(i)}, S_t^{b(i)})}
\end{equation}

\subsection{Neural Statistics-Level Consistency} 
\label{subsec:statistics-level_consistency}
As pointed out in \cite{li2021momentum, cai2021exponential} that the mismatch in batch normalization statistics between teacher and student models could lead to suboptimal model performance, in our case, while the student model takes both source domain data $x_s$ and target domain data $\hat{x_t}$ as input, the teacher model only has access to the target data $x_t$. The distribution shift lying between source and target data could lead to mismatched batch statistics between the batch normalization (BN) layers of the student and teacher models. This mismatch could cause misaligned normalization and in turn, leads to an unstable training process with degraded performance or even divergence. We provide an in-depth analysis regarding this matter in Section \ref{subsec:further_analysis}.

To mitigate this issue, we propose to use the running statistics of the student model BN layers for the teacher model during the training process. Specifically, for each BN layer in the student model, the batch mean $\mu$ and variance $\sigma$ are used to update the running statistics at every iteration:
\begin{align}
     \mu' = (1-\alpha) \mu' + \alpha \mu \\
     \sigma' = (1-\alpha) \sigma' + \alpha \sigma
\end{align}
where $\mu'$ and $\sigma'$ are the running mean of $\mu$ and $\sigma$ and $\alpha$ is the BN momentum that controls the speed of batch statistics updating the running statistics. For the teacher model, we use $\mu'$ and $\sigma'$ instead of the batch statistics for all the BN layers to normalize the layer inputs. We argue that this modification closes the gap caused by domain mismatch and leads to more stable training behavior. We empirically demonstrate the effectiveness by comparing the performance under different BN settings in Section \ref{subsec:ablation_study}.

\begin{table*}[t]
    \centering
    \small
    \caption{Performance of \name{} on four source-target pairs in comparison with various baselines and state-of-the-art methods. \name{} outperforms all baselines and even surpasses SOTA methods that utilize target domain annotation information (indicated by $\dagger$). Direct transfer: the model trained on the source domain is directly tested on the target domain. Wide-Range Aug: baseline method with random scaling augmentation of a wide range which potentially includes the target domain scales. It is thus validated the drastic performance degradation cannot be fully mitigated by simple data augmentation. DA-Faster\cite{chen2018dafastercnn}: a representative method based on adversarial feature alignment, a common technique used in 2D domain adaptation. $\#$ indicates the implementation is adapted from 2D to 3D. However, feature alignment is unable to solve the geometric mismatch, which we argue is unique to 3D detection. The state-of-the-art work \cite{wang2020traininger} proposes to perform output transformation (OT) to scale predictions and statistical normalization (SN) for scale-adjusted training examples. Both OT and SN require known target domain statistics. \name{}, albeit being fully unsupervised, even surpasses these methods on key metrics: APH/L2 (Waymo), AP$^{3D}$ (nuScenes), and AP Moderate (KITTI).}
    \begin{tabular}{c|c|c|c|c"c|c|c|c}
        \thickhline
        \multicolumn{5}{c"}{\textbf{KITTI $\rightarrow$ Waymo}} & \multicolumn{4}{c}{\textbf{Waymo $\rightarrow$ KITTI}} \\
        \thickhline
        Methods & AP/L1 & APH/L1 & AP/L2 & APH/L2 & 
        Methods & Easy &  Moderate & Hard \\
        \hline
        Direct Transfer & 9.17 &	8.99 & 7.94 & 7.78 & 
        Direct Transfer & 20.22 & 21.43 & 20.49 \\
        Wide-Range Aug & 18.61 & 18.18 & 16.77 & 16.40 & 
        Wide-Range Aug & 30.23 & 31.49 & 32.85 \\
        DA-Faster~\cite{chen2018dafastercnn}$^{\#}$ &  6.96 &	6.87 & 6.42 & 6.33 &
        DA-Faster~\cite{chen2018dafastercnn}$^{\#}$ & 4.42 & 5.55 & 5.53	\\
        OT \cite{wang2020traininger}$^{\dagger}$ & 26.48 &	25.84 & 23.85	& 23.29 & 
        OT \cite{wang2020traininger}$^{\dagger}$ & 39.78 & 37.82 & 39.55 \\
        SN \cite{wang2020traininger}$^{\dagger}$ & 30.69 &	30.06 & 27.23 & 26.67 &
        SN \cite{wang2020traininger}$^{\dagger}$ & 61.93 & 58.07 & \textbf{58.44} \\
        Ours & \textbf{38.21} & \textbf{37.74} & \textbf{34.46} & \textbf{34.04} & 
        Ours & \textbf{69.35} & \textbf{59.44} & 56.29 \\
        
        \thickhline
        \multicolumn{5}{c"}{\textbf{KITTI $\rightarrow$ nuScenes}} & \multicolumn{4}{c}{\textbf{nuScenes $\rightarrow$ KITTI}} \\
        \thickhline
        Methods & ATE & ASE & AOE & AP$^{3D}$ & 
        Methods & Easy &  Moderate & Hard  \\
        \hline
        Direct Transfer  &  0.207 &  0.248 &  0.212 &  13.01 &
        Direct Transfer  & 49.13 &  39.56 & 35.51 \\
        Wide-Range Aug & 0.200  & 0.228 &  0.211 & 16.01 & 
        Wide-Range Aug  & 58.71 &  45.37 &  43.03 \\
        DA-Faster~\cite{chen2018dafastercnn}$^{\#}$ & 0.247  & 0.253  & 0.292  & 10.77 &
        DA-Faster~\cite{chen2018dafastercnn}$^{\#}$ & 52.25 & 40.62 & 35.90 \\
        OT \cite{wang2020traininger}$^{\dagger}$ & 0.207  &   0.220  &   0.212  & 14.67 &
        OT \cite{wang2020traininger}$^{\dagger}$ &  23.13 & 	27.26 & 	29.10 \\
        SN \cite{wang2020traininger}$^{\dagger}$ & 0.227 &  \textbf{0.168} &  0.368 &  23.15 &
        SN \cite{wang2020traininger}$^{\dagger}$ &  44.81 & 	45.15	 & 47.60 \\
        Ours  & \textbf{0.197}  &  0.179  &  \textbf{0.197} & \textbf{23.47}  & 
        Ours  & \textbf{71.26} &  \textbf{55.42} & \textbf{48.99} \\
        \thickhline
    \end{tabular}
    \label{tab:benchmark}
    \vspace{-0.5em}
\end{table*}

\section{Experiments} 
\label{sec:exp}
We first introduce the popular autonomous driving datasets including KITTI \cite{geiger2013kittidataset}, Waymo Open Dataset \cite{sun2020waymo}, and nuScenes \cite{caesar2020nuscenes} used in the experiments (Section~\ref{subsec:datasets}). We then benchmark \name{} across datasets where \name{} achieves consistent performance boost in Section~\ref{subsec:benchmarking_results}. Moreover, we ablate \name{} to give a comprehensive assessment of its submodules and justify our design choices in Section~\ref{subsec:ablation_study}. Finally, we further investigate the challenges of unsupervised domain adaptive 3D detection and show \name{} successfully addresses them. We further analyse the problems in 3D domain adaptive detection and our solutions in Section~\ref{subsec:further_analysis}. Due to space constraint, we include the implementation details in the Supplementary Material.

\subsection{Datasets}
\label{subsec:datasets}
We follow \cite{wang2020traininger} to evaluate \name{} on various source-target combinations with the following datasets. 

\noindent \textbf{KITTI.} KITTI \cite{geiger2013kittidataset} is a popular autonomous driving dataset that consists of 3,712 training samples and 3,769 validation samples. The 3D bounding box annotations are only provided for objects within the Field of View (FoV) of the front camera. Therefore, points outside of the FoV are ignored during training and evaluation. We use the official KITTI evaluation metrics for evaluation where the objects are categorized into three levels (Easy, Moderate, and Hard) and the mean average precision is evaluated.

\noindent \textbf{Waymo Open Dataset.} The Waymo Open Dataset (referred to as Waymo) \cite{sun2020waymo} is a large-scale benchmark that contains 122,000 training samples and 30,407 validation samples. We subsample 1/10 of the training and validation set. To align the input convention, we apply the same front camera FoV as the KITTI dataset. The official Waymo evaluation metrics including mean average precision (AP) and mean average precision weighted by heading (APH) are used to benchmark the performance for objects of two difficulty levels (L1 and L2).

\noindent \textbf{nuScenes.} The nuScenes \cite{caesar2020nuscenes} dataset consists of 28,130 training samples and 6,019 validation samples. We subsample the training dataset by 50\% and use the entire validation set. We also apply the same FoV on the input as other datasets. We adopt the official evaluation metrics of translation, scale, and orientation errors, with the addition of the commonly used average precision based on 3D IoU with a threshold of 0.7 to reflect the overall detection accuracy.



\subsection{Benchmarking Results}
\label{subsec:benchmarking_results}
As an emerging research area, the cross-domain point cloud detection topic has relatively small literature. To the best of our knowledge, \cite{wang2020traininger} is the most relevant work that has a similar setting as our study. We compare our method with two normalization methods proposed in \cite{wang2020traininger}, namely Output Transformation (OT) and Statistical Normalization (SN), where the former transforms the predictions by an offset and the latter trains the detector with scale-normalized input. Moreover, we also compare with the adversarial feature alignment method, which is commonly used on image-based tasks, by adapting DA-Faster \cite{chen2018dafastercnn} to our PointRCNN \cite{shi2019pointrcnn} base model. We also provide Direct Transfer and Wide-Range Augmentation as baselines. Figure \ref{fig:teaser} displays a qualitative comparison of the detection results before and after domain adaptation with our proposed method. More results can be found in the Supplementary Material.

Table \ref{tab:benchmark} demonstrates the cross-domain detection performance on four source-target domain pairs, \name{} outperforms all unsupervised baselines by convincing margins. We highlight that our method adapts scale for each instance instead of applying a global shift, allowing us to surpass state-of-the-art methods that utilize target domain object scale statistics.

\subsection{Ablation Study}
\label{subsec:ablation_study}

To evaluate the effectiveness of the components of \name{}, we conduct ablation studies on KITTI $\to$ Waymo transfer with PointRCNN as the base model.

\smallskip
\noindent\textbf{Effectiveness of Point/Instance-Level Consistency.}
We study the effects of different components of the proposed consistency loss. Table \ref{tab:ablation} reports the experimental results when different combinations of loss components are applied. It is observed that for both point-level consistency and instance-level consistency, the box consistency clearly has a larger contribution as compared to the class consistency. This observation indicates that the scale difference is a major source of the domain gap between source and target domains with different object size distributions, which is also in line with the previous work \cite{wang2020traininger}. It also shows that our proposed box consistency regularization method effectively mitigates this gap. In addition, all losses are complementary to one another: the best result is achieved when all four of them are used.

\begin{table}[t]
  \centering
  \fontsize{11}{12}\selectfont
    \caption{Ablation study of point-level and instance-level consistency loss components. Results show loss components are highly complementary; the joint use of all four losses at two levels achieves the best performance. More importantly, we find that the bounding box regression loss, which is directly associated with bounding box scale, benefits the performance more than the classification loss. This further validates our stance that geometric mismatch is a key domain gap for 3D detection.}
  \scalebox{0.7}{
  \begin{tabular}{p{0.8cm}<{\centering}p{0.8cm}<{\centering}p{0.8cm}<{\centering}p{1.1cm}<{\centering}|cccc}
    \thickhline
    $L_{pt,cls}$ & $L_{pt,box}$ & $L_{ins,cls}$ & $L_{ins,box}$ & AP/L1 & APH/L1 & AP/L2 & APH/L2 \\
    \hline
    & & & & 18.61 &	18.18 &	16.77	 & 16.40 \\
    $\checkmark$ & & & & 20.34  & 	19.91	 & 18.07 & 	17.70 \\
     & $\checkmark$ & & & 30.34 & 	29.69 & 	27.08	 & 26.49 \\
    $\checkmark$ & $\checkmark$ & & & 31.00 &	30.39 &	27.64 &	27.09 \\
    & & $\checkmark$ & & 21.12 & 	20.87	 & 18.79 & 	18.57 \\
    & &  & $\checkmark$ & 33.21	 & 32.44 & 	29.95 & 	29.26 \\
    & & $\checkmark$ & $\checkmark$ & 34.95 & 	34.53 & 	31.43	 & 31.05 \\
    $\checkmark$ & $\checkmark$ & $\checkmark$ & $\checkmark$ & \textbf{38.21}	 & \textbf{37.74} & \textbf{34.46} & \textbf{34.04} \\
    \thickhline
  \end{tabular}
  }
  \label{tab:ablation}
  \vspace{-0.5em}
\end{table}

Furthermore, we compare \name{} with two alternative approaches for point and box matching respectively in Table \ref{tab:matching_methods}. Compared to these baseline approaches, \name{} replicates the input point clouds and the region proposals before they are passed to the student and teacher models to eradicate any noise which may arise from inaccurate matching. The results highlight the importance of correspondence in constructing meaningful consistency losses for effective unsupervised learning.

\begin{table}[t]
    \centering
    \caption{Ablation study of point-level and instance-level matching methods. Nearest Point: a baseline for point match where a point in the student input is matched to the nearest point in the teacher input using Euclidean distance. Max IoU Box: a baseline for box matching where a student box prediction is matched to the teacher pseudo label with the largest IoU. Ours: input point clouds or region proposals of the student are replicated from the teacher. We highlight that our matching method ensures accurate one-to-one correspondence, which is critical to effective teacher-student learning.}
    \scalebox{0.85}{
    \begin{tabular}{c|cccc}
        \thickhline
        Matching Method & AP/L1 & APH/L1 & AP/L2 & APH/L2 \\
        \hline
        Nearest Point &  2.93 & 2.86 & 2.65 & 2.58  \\
        Max IoU Box    & 26.95 & 26.66 & 24.18 & 23.92 \\
        Ours  & \textbf{38.21} & \textbf{37.74} & \textbf{34.46} & \textbf{34.04}   \\
        
        \thickhline
    \end{tabular}}
    \label{tab:matching_methods}
    \vspace{-0.5em}
\end{table}

\noindent\textbf{Effectiveness of Neural Statistics-Level Consistency.}
We also experiment on the effectiveness of neural statistics-level consistency by comparing the performance when such alignment is enabled and disabled. From Table \ref{tab:bn} we can see that when neural statistics-level consistency is disabled, the model performance severely drops. As analyzed in Section \ref{subsec:statistics-level_consistency}, when neural statistics-level consistency is not in place, the teacher model BN layers normalize the input features using batch statistics that are obtained from only target data, while the student model performs BN with statistics from both source and target domains. This misalignment creates a significant gap. As a result, the consistency computation between the student and teacher predictions is invalidated. We also compare with the approach that the student model performs separate BN for source and target data. In this case, although the normalization for target input is performed with target statistics for both models, the mismatched normalization of source and target inputs leads to suboptimal performance as compared to \name{}.

\begin{table}[t]
    \centering
    \caption{Ablation study of neural statistics-level consistency indicates that \name{} effectively closes the domain gap due to neural statistics mismatch. Disabled: no consistency is enforced. Separate: the student model performs BN separately for source and target domain inputs to align with the teacher model. Enabled: our proposed neural statistics-level alignment.}
    \scalebox{0.85}{
    \begin{tabular}{c|cccc}
        \thickhline
        Setting  & AP/L1 & APH/L1 & AP/L2 & APH/L2 \\
        \hline
        Disabled & 2.79 &	2.74 & 2.54 & 2.49 \\
        Separate & 29.88 &	29.45 & 26.85 & 26.48 \\
        Enabled  & \textbf{38.21} & \textbf{37.74} & \textbf{34.46} & \textbf{34.04} \\
        \thickhline
    \end{tabular}
    }
    \label{tab:bn}
     \vspace{-0.5em}
\end{table}

\noindent\textbf{Effectiveness of Mean Teacher.}
The teacher model is essentially a temporal ensemble of student models at different time stamps. We study the effectiveness of the mean teacher paradigm by comparing the performance when the exponential moving average update is enabled or disabled. Table \ref{tab:meanteacher} shows that it is important to employ the moving average update mechanism for the teacher to generate meaningful supervisions to guide the student model, and the removal of such mechanism leads to performance deterioration.

\begin{table}[t]
    \centering
    \caption{Ablation study of the exponential moving average (EMA) update scheme in mean teacher paradigm. The performance significantly degrades when the exponential moving average update is disabled, highlighting the importance of the mean teacher design in producing meaningful targets.}
    \scalebox{0.85}{
    \begin{tabular}{c|cccc}
    \thickhline
    EMA & AP/L1 & APH/L1 & AP/L2 & APH/L2  \\
    \hline
    Disabled & 8.95 & 	8.66	 & 8.35	 & 8.08 \\
    Enabled  & \textbf{38.21} & \textbf{37.74} & \textbf{34.46} & \textbf{34.04} \\
    \thickhline
    \end{tabular}
    }
    \label{tab:meanteacher}
    \vspace{-0.5em}
\end{table}

\subsection{Further Analysis}
\label{subsec:further_analysis}

\noindent\textbf{Analysis of Distribution Shift.}
%
We highlight that the geometric mismatch is a significant issue for cross-domain deployment of 3D detection models. In Figure \ref{fig:distribution}, the object dimension (length, width, and height) distributions are drastically different across domains with a relatively small overlap. The baseline, trained on the source domain, is not able to generalize to the target domain as the distribution of its dimension prediction is still close to that of the source domain. In contrast, \name{} is able to adapt to the new domain by predicting highly similar geometric distribution as the target domain. 

\begin{figure}[t]
    \centering
    \includegraphics[width=0.9\linewidth]{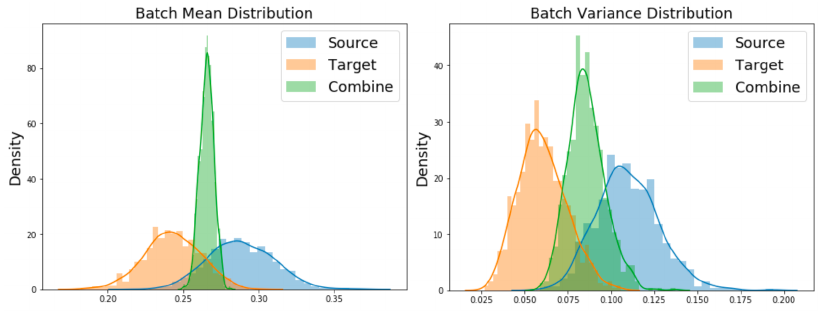}
    \caption{Neural statistics mismatch across domains. We plot the distributions of batch mean and batch variance. Significant misalignment in batch statistics between source and target domains is observed, which highlights the necessity of neural statistics-level consistency.}
    \label{fig:bn_distribution}
    \vspace{-0.5em}
\end{figure}

\noindent\textbf{Analysis of Neural Statistics Mismatch.} Figure \ref{fig:bn_distribution} shows that inputs from different domains have very different distributions of batch statistics, which explains the tremendous performance drop when our proposed neural statistics-level consistency is not applied to align the statistics (Table \ref{tab:bn}).

\begin{figure}
    \centering
    \includegraphics[width=0.9\linewidth]{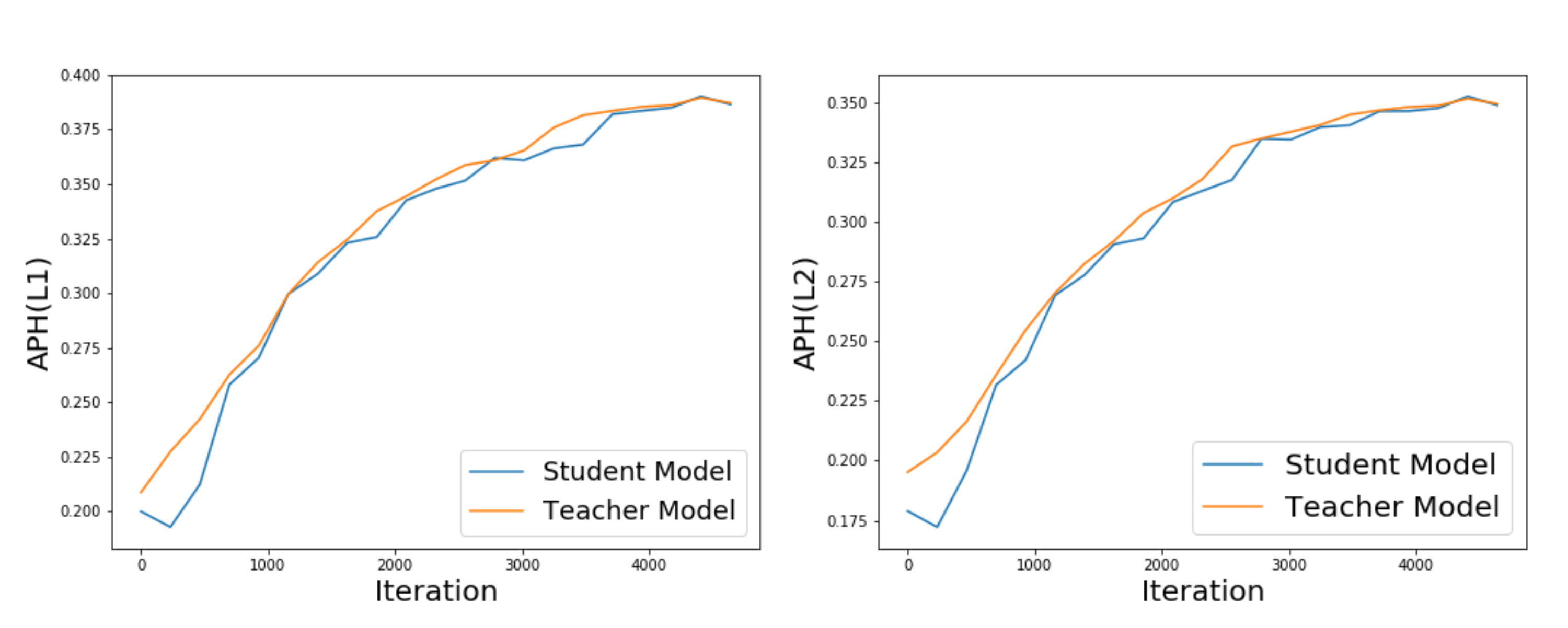}
    \caption{Teacher and student model performance against iteration. Not only does the teacher model constantly outperform the student, its performance curve is also smoother. Hence, the teacher model, which can be regarded as a temporal ensemble of the student model, is able to produce more stable and accurate pseudo labels to supervise the student model.}
    \label{fig:student_vs_teacher}
    \vspace{-0.5em}
\end{figure}

\noindent\textbf{Analysis of Teacher/Student Paradigm.}
In Figure \ref{fig:student_vs_teacher}, the teacher model in \name{} demonstrates stronger performance during the training process until convergence. Moreover, the teacher model exhibits a smoother learning curve. This validates the effectiveness of our mean-teacher paradigm to create accurate and reliable supervision for robust optimization of the student model.

\section{Conclusion}
\label{sec:conclusion}
We study unsupervised 3D domain adaptive detection that requires no target domain annotation or annotation-related statistics. We validate that geometric mismatch is a major contributor to the domain shift and propose \name{} that leverages a teacher-student paradigm for robust and reliable pseudo label generation via point-, instance- and neural statistics-level consistency to enforce effective transfer. \name{} outperforms all the baselines by convincing margins, and even surpasses methods that require additional target information.

\smallskip
\noindent \textbf{Acknowledgements} This study is supported by NTU NAP, and under the RIE2020 Industry Alignment Fund – Industry Collaboration Projects (IAF-ICP) Funding Initiative, as well as cash and in-kind contribution from the industry partner(s).

\clearpage
{\small
\bibliographystyle{ieee_fullname}
\bibliography{mybibfile}
}
\clearpage

\section{Supplementary Materials}

\subsection{Overview}
We provide additional implementation details, experiment results and visualizations that are not included in the main paper due to space constraint. 

\begin{itemize}
    \item Section \ref{3dssd_implementation}. We describe the implementation details and the experimental results of MLC-Net with 3DSSD \cite{yang20203dssd} used as the base model.
    \item Section \ref{sec:implementation_details}. We provide the implementation details of our MLC-Net (PointRCNN \cite{shi2019pointrcnn} as the base model) and other baselines that are discussed in the main paper.
    \item Section \ref{sec:additional_experiment}. Additional experimental results.
    \item Section \ref{sec:qualitative}. More qualitative visualization results.
\end{itemize}

\subsection{Implementation based on 3DSSD}
\label{3dssd_implementation}
We demonstrate that our proposed method is detector-agnostic by adapting MLC-Net to one-stage detector 3DSSD. 

\subsubsection{MLC-Net on 3DSSD}
\begin{figure}[h]
    \centering
    \includegraphics[width=1.0\linewidth]{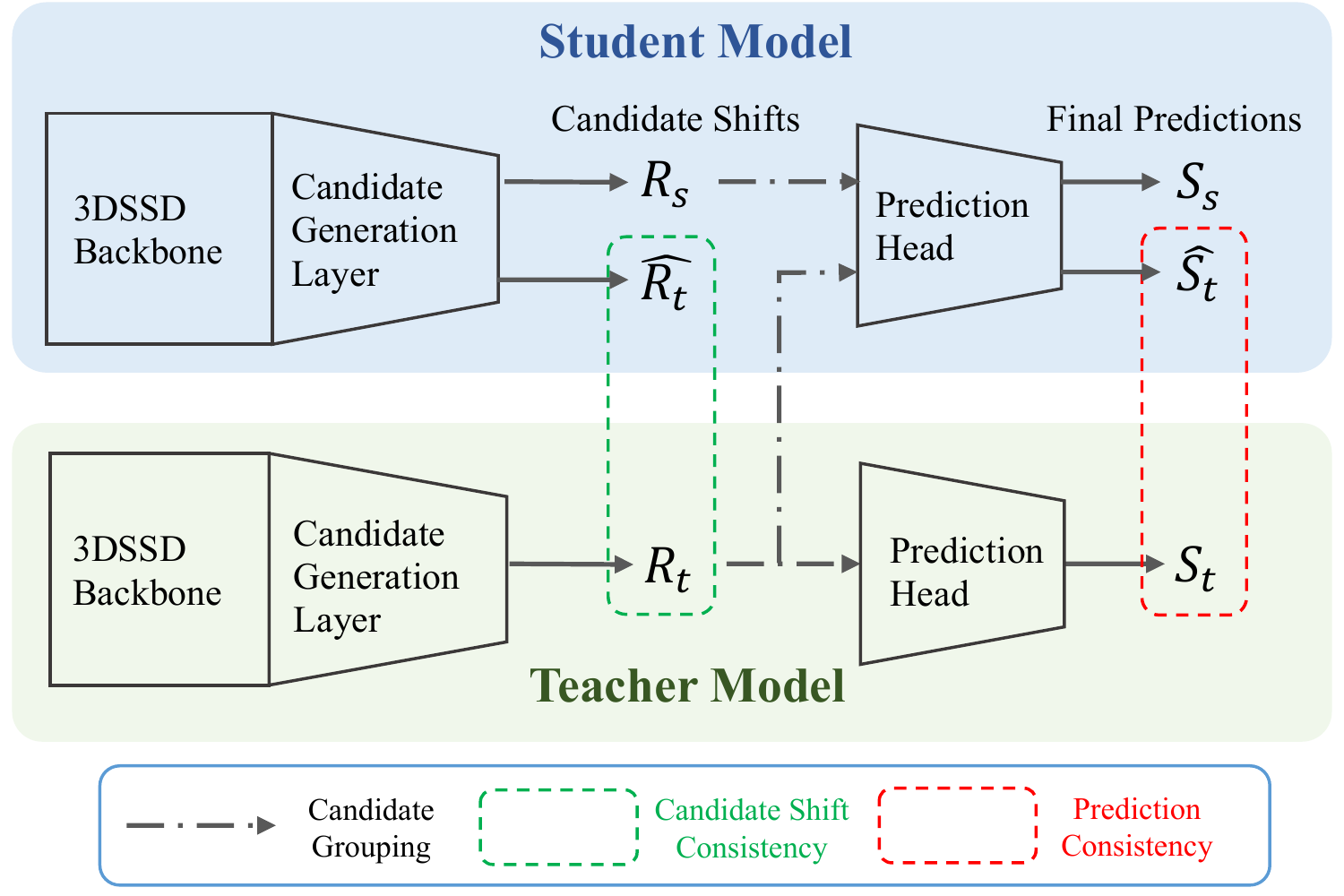}
    \caption{Architecture of MLC-Net with 3DSSD as the base model. The consistency loss is computed based on the candidates shifts $R$ and the final predictions $S$. Candidate points generated by the teacher model is passed to the student model to establish correspondence between student prediction $\hat{S}_t$ and teacher prediction $S_t$.}
    \label{fig:3dssd}
\end{figure}

Being a one-stage detector, 3DSSD differs from two-stage detectors that it does not generate region proposals. Instead, as shown in Figure \ref{fig:3dssd}, 3DSSD first employs a modified PointNet++ \cite{qi2017pointnet++} model to extract point cloud features and downsample the points. A candidate generation layer is then applied to predict candidate shifts $R$ which are the offsets of object locations relative to the downsampled points. The corrected points are treated as candidate points and candidate grouping is performed to generate instance-level features, and the final predictions $S$ are predicted by the prediction head. Please refer to \cite{yang20203dssd} for more details. 
 
Despite that 3DSSD does not have the region proposal stage as PointRCNN, we highlight that MLC-Net is highly compatible as long as there are point-based operations and final instance predictions. Catering to the model architecture, we compute the consistency loss based on the difference between the student and teacher predictions of $R$ and $S$. For candidate shifts $R$, we establish the point correspondences by passing the sampling index of the teacher model to the student model. As a result, both models sample the same points and the one-to-one matching of points is established. Similar to our implementation based on PointRCNN where region proposals of the teacher model is used for feature pooling at the student model, we copy the candidate points from the teacher to the student model for candidate grouping to obtain instance-level features. This operation guarantees the correspondences of final predictions $S$. The rest of the operations are similar to that of our method introduced in the main paper.

\subsubsection{Experiments}
We evaluate MLC-Net implemented with 3DSSD on KITTI \cite{geiger2013kittidataset} and Waymo \cite{sun2020waymo} datasets. For KITTI to Waymo transfer, we pretrain the model on KITTI dataset for 80 epochs and finetune with our proposed method for 20 epochs. For Waymo to KITTI transfer, we pretrain the model on Waymo dataset for 40 epochs and finetune with MLC-Net for 5 epochs. Note that all the source domain examples are used once in each training epoch with the target domain data randomly sampled or resampled to match the number. The number of epochs for the Waymo to KITTI transfer task is lower because the Waymo dataset has a substantial larger number of samples as compared to KITTI. During finetuning, we use the ADAM \cite{kingma2014adam} optimizer with a learning rate of 0.001. The batch size is set to 32.

Table \ref{tab:ssdk2w} and Table \ref{tab:ssdw2k} report the performance comparison for KITTI to Waymo and Waymo to KITTI transfer tasks, respectively. It can be seen that MLC-Net consistently outperforms the Statistical Normalization (SN) method for both tasks. It is interesting to observe that the cross-domain detection performance of 3DSSD is lower than PointRCNN on these tasks, while 3DSSD has stronger in-domain performance as reported in \cite{yang20203dssd}. This could be attributed to the one-stage design of 3DSSD that no bounding box refinement is performed, which makes it less robust to scale variations.

\begin{table}[t]
    \centering
    \caption{Performance comparison on Waymo validation dataset for transfer from KITTI to Waymo with 3DSSD as the base model.}
    \scalebox{0.85}{
    \begin{tabular}{c|cccc}
        \thickhline
        \multicolumn{5}{c}{\textbf{KITTI $\rightarrow$ Waymo}} \\
        \thickhline
        Methods & AP/L1 & APH/L1 & AP/L2 & APH/L2  \\
        \hline
        Direct Transfer & 3.29 &	3.26 & 2.78 & 2.75 \\
        Wide-Range Aug & 16.67 & 16.48 & 14.73 & 14.56 \\
        OT \cite{wang2020traininger} & 24.56 &	24.23 & 22.70	& 22.39 \\
        SN \cite{wang2020traininger} & 25.95 &	25.61 & 24.00 & 23.67 \\
        Ours & \textbf{29.87} & \textbf{29.27} & \textbf{26.80} & \textbf{26.27} \\ 
        \thickhline
    \end{tabular}
    }
    \label{tab:ssdk2w}
\end{table}

\begin{table}[t]
    \centering
    \caption{Performance comparison on KITTI validation dataset for transfer from Waymo to KITTI with 3DSSD as the base model.}
    \scalebox{0.85}{
    \begin{tabular}{c|ccc}
        \thickhline
        \multicolumn{4}{c}{\textbf{Waymo $\rightarrow$ KITTI}} \\
        \thickhline
        Methods & Easy & Moderate & Hard \\
        \hline
        Direct Transfer & 6.31 &	6.41 & 6.25 \\
        Wide-Range Aug & 37.83 & 35.34 & 34.05 \\
        OT \cite{wang2020traininger} & 45.42 & 40.50 & 41.04 \\
        SN \cite{wang2020traininger} & 47.81 & 45.92 & 46.46 \\
        Ours & \textbf{56.86} & \textbf{48.74} & \textbf{48.32} \\ 
        \thickhline
    \end{tabular}
    }
    \label{tab:ssdw2k}
\end{table}

\subsection{Implementation Details}
\label{sec:implementation_details}
In this section, we provide implementation details of MLC-Net with PointRCNN as the base model as well as other baselines used in comparison.

\subsubsection{MLC-Net on PointRCNN}
For our PointRCNN-based implementation of MLC-Net discussed in the main paper, we build our method on the official implementation \cite{openpcdet2020}. For all the experiments, we first pretrain the base model with source data and load the student and teacher models with the same pretrained weights as initialization. For the pretraining, the default configurations provided in \cite{openpcdet2020} are used. We then train the models with our proposed method with ADAM optimizer and a learning rate of 0.0001. A batch size of 32 is used. When KITTI is the source domain, we train the MLC-Net for 20 epochs and set $m$ as 0.99 and $\alpha$ as 0.05. When transferring to KITTI, we conduct training for 5 epochs set $m$ and $\alpha$ as 0.999 and 0.001, respectively. For all the experiments, the source domain random scaling augmentation range is set to [0.7, 1.3], while the target domain input augmentation $h$ and RoI augmentation $\xi$ both use a range of [0.9, 1.1]. The probability threshold $\varepsilon$ is set to 0.99 and the loss weight coefficient $\lambda$ is set to 0.1. We follow \cite{wang2020traininger} and conduct all the evaluations on the car category. We implement our method using Pytorch and run the experiments with 8 NVIDIA V100 GPUs.

\subsubsection{Other Baselines}
For the comparing methods in Section \textcolor{red}{4.2} of the main paper, Direct Transfer uses a default random scaling input augmentation range of [0.95, 1.05], while Wide-Range Aug refers to a wide random scaling input augmentation range of [0.7, 1.3], which is the same as the setting for MLC-Net. To adapt DA-Faster \cite{chen2018dafastercnn} to PointRCNN, we apply two domain discriminators to align the feature representations. One of the discriminators is applied to the global features obtained from the PointNet++ \cite{qi2017pointnet++} backbone, while the other discriminator is applied to the instance-level features obtained from point cloud region pooling. 


\subsection{Additional Experiment Results}
\label{sec:additional_experiment}

\noindent\textbf{Effectiveness of Augmentations.}
Moreover, we evaluate the significance of input augmentation $h$ and RoI augmentation $\xi$ in Table~\ref{tab:augmentation}. The use of augmentations consistently improve the performance of the model as random perturbations force the network to adapt to a wide range of distributions. We highlight that data augmentation is able to further boost the performance of MLC-Net, which already outperforms all baselines and state-of-the-art methods without any augmentation.

\begin{table}[t]
    \centering
    \caption{Ablation study of random data augmentation. Both input and RoI augmentations force the model to be more adaptive to scale variations, and are found to be useful. They are also complementary to each other: applying both achieves the best result.}
    \scalebox{0.85}{
    \begin{tabular}{cc|cccc}
        \thickhline
        Input Aug & RoI Aug & AP/L1 & APH/L1 & AP/L2 & APH/L2 \\
        \hline
                   &            & 34.34 & 33.89 & 30.86 & 30.45 \\
        \checkmark &            & 36.23 & 35.81 &	 32.57 & 32.19 \\
                   & \checkmark & 35.86 & 35.44 & 32.22 & 31.85 \\
        \checkmark & \checkmark & \textbf{38.21} & \textbf{37.74} & \textbf{34.46} & \textbf{34.04} \\
        \thickhline
    \end{tabular}}
    \label{tab:augmentation}
\end{table}

\smallskip
\noindent\textbf{Additional Comparison with SF-UDA\textsuperscript{3D} \cite{saltori2020sourcefree}.}
As introduced in the related works, SF-UDA\textsuperscript{3D} proposes to address the 3D domain adaptive detection problem by leveraging the temporal coherence of target predictions. Specifically, the model trained on the source domain is used to generate predictions given target domain inputs of different scales. The best scale is selected by comparing the prediction consistency over a number of consecutive frames. Subsequently, target predictions of the best scale are used to finetune the pretrained model. SF-UDA\textsuperscript{3D} requires consecutive point cloud frames as the input, which is not directly comparable to our proposed method which only requires single-frame input. Nevertheless, we compare the performance on the nuScenes $\to$ KITTI transfer task where the same evaluation metrics are used for both methods with the same PointRCNN base model. Table \ref{tab:sfuda} shows that our proposed MLC-Net outputs SF-UDA\textsuperscript{3D} without leveraging any temporal information.

\begin{table}[t]
    \centering
    \caption{Comparision with SF-UDA\textsuperscript{3D} on nuScenes $\to$ KITTI transfer on the base model of PointRCNN. $^{*}$ indicates the results are reprinted from the original paper. Our MLC-Net outperforms SF-UDA\textsuperscript{3D} without any requirement for temporal information.}
    \scalebox{0.85}{
    \begin{tabular}{c|c|ccc}
    \thickhline
        Methods & Require Sequence & Easy & Moderate & Hard \\
        \hline
        Direct Transfer  & & 49.1 &  39.6 & 35.5 \\
        SF-UDA\textsuperscript{3D} \cite{saltori2020sourcefree}$^{*}$ & \checkmark & 68.8 & 49.8 & 45.0  \\
        Ours & & \textbf{71.3} &  \textbf{55.4} & \textbf{49.0} \\
    \thickhline
    \end{tabular}
    }
    \label{tab:sfuda}
\end{table}

\subsection{Qualitative Results}
\label{sec:qualitative}
\begin{figure*}[t]
    \centering
    \includegraphics[width=1.0\linewidth]{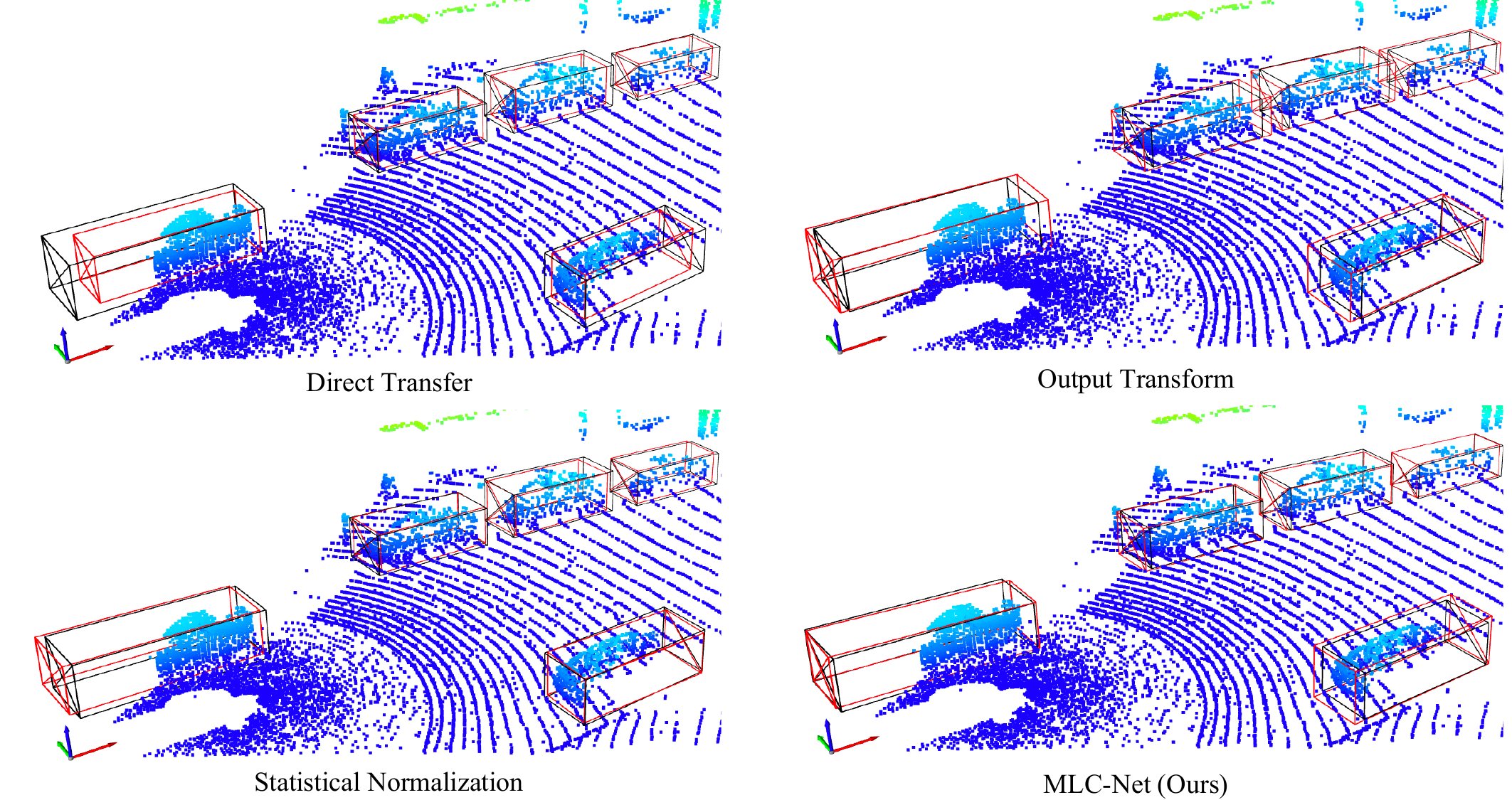}
    \caption{Qualitative results on Waymo validation dataset for KITTI to Waymo transfer.}
    \label{fig:k2w_visual}
\end{figure*}

\begin{figure*}[t]
    \centering
    \includegraphics[width=1.0\linewidth]{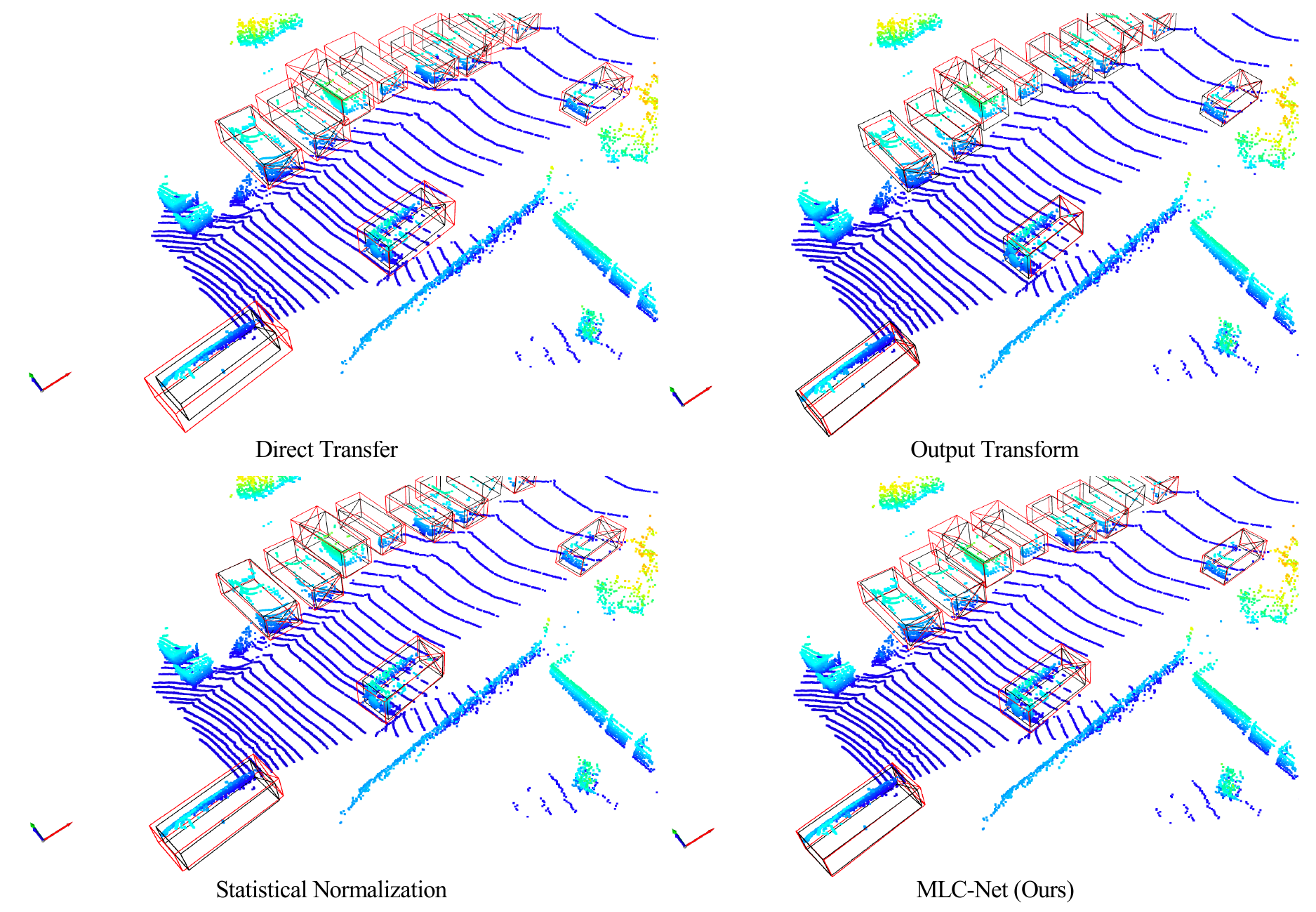}
    \caption{Qualitative results on KITTI validation dataset for Waymo to KITTI transfer.}
    \label{fig:w2k_visual}
\end{figure*}

As shown in Figure \ref{fig:k2w_visual} and Figure \ref{fig:w2k_visual} ,we provide additional visualization of cross-domain detection results of different methods on multiple transfer tasks. It can be observed that directly applying a model trained on the source domain (Direct Transfer) suffers from significantly degraded performance due to geometric mismatches. While all the domain adaptation approaches demonstrate effectiveness in correcting the scale, Output Transform often under-corrects or over-corrects the predictions and cause inaccurate localization. This can be attributed to the global offset applied to all the bounding boxes, whereas individual predictions require different corrections. MLC-Net is able to mitigate the geometric mismatches effectively on various transfer tasks, which demonstrates the domain adaption capability of our proposed method.

\clearpage

\end{document}